# Online Dynamic Motion Planning and Control for Wheeled Biped Robots

Songyan Xin and Sethu Vijayakumar

*Abstract*—Wheeled-legged robots combine the efficiency of wheeled robots when driving on suitably flat surfaces and versatility of legged robots when stepping over or around obstacles. This paper introduces a planning and control framework to realise dynamic locomotion for wheeled biped robots. We propose the Cart-Linear Inverted Pendulum Model (Cart-LIPM) as a template model for the rolling motion and the under-actuated LIPM for contact changes while walking. The generated motion is then tracked by an inverse dynamic whole-body controller which coordinates all joints, including the wheels. The framework has a hierarchical structure and is implemented in a model predictive control (MPC) fashion. To validate the proposed approach for hybrid motion generation, two scenarios involving different types of obstacles are designed in simulation. To the best of our knowledge, this is the first time that such online dynamic hybrid locomotion has been demonstrated on wheeled biped robots.

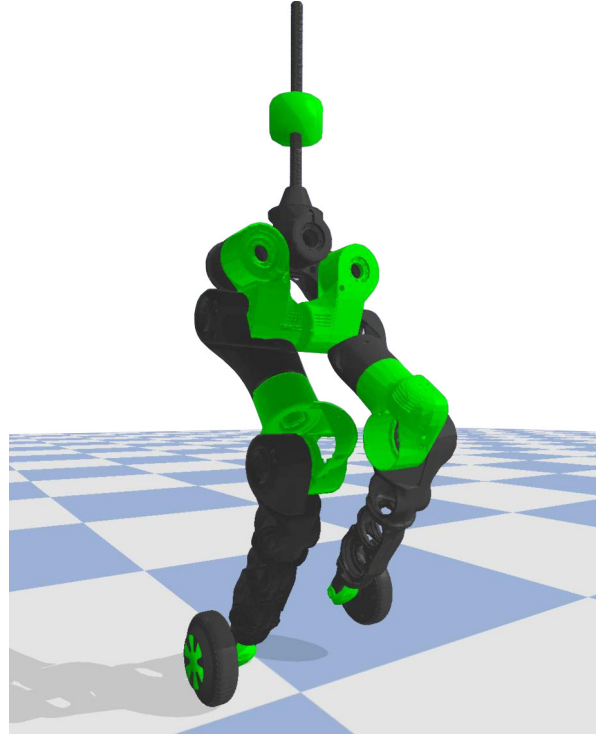

Fig. 1: Wheeled biped robot in hybrid locomotion.

## I. INTRODUCTION

Wheeled robots move faster and more efficiently than legged robots in a structured environment. However, legged robots are more capable at traversing challenging terrains such as stairs and narrow trenches. Wheeled-legged robots have the potential to combine the best of both worlds. In this paper, we will focus on hybrid locomotion for wheeled biped robots. Hybrid locomotion refers to simultaneous rolling and walking motion as shown in Fig. 1. It can help the robot stepping over or around obstacles. Such motion is difficult to realise on wheeled biped robots since the robot needs to balance in both forward and lateral directions simultaneously. The balancing problem in the forward direction has been studied extensively on two-wheeled self-balancing robots. The balancing issue in the lateral direction is actually a walking problem and existing studies on biped robots can provide a lot of insight. We unify the two in this paper to realise hybrid locomotion.

### A. Literature Review

For two-wheeled robots, the control problem mainly focuses on balancing in the sagittal plane (pitch motion) [1]. The most commonly used template model is the Wheeled Inverted Pendulum Model (WIPM) [2]. Linearization of this non-linear model around its upright stable equilibrium configuration is needed to enable linear state feedback control [3]. Alternatively, a non-linear approach can be directly applied. In [4] differential dynamic programming (DDP) and model predictive control (MPC) are applied to generate whole body motion for a wheeled humanoid robot. However, the computation time is not given. In this paper, we propose the new Cart-LIPM model for controlling the motion in this direction. It remains linear in a large range of motion compared to the WIPM linearized around a fixed point. Due to its linearity, it enables a much higher MPC update frequency comparing to the non-linear MPC.

LIPM has been widely used to generate walking motion for biped robots [5] [6]. An important consideration when using this model for walking motion generation is the actuation type associated with the foot. Fully-actuated LIPM assumes planar feet and ankle torques actuation, as a result, the zero moment point (ZMP) can be modulated inside the supporting area. In contrast, the under-actuated LIPM assumes point feet and therefore it has no insole ZMP modulation

The authors are with School of Informatics, University of Edinburgh, Edinburgh, EH8 9AB, United Kingdom. Email: sxin@ed.ac.uk, sethu.vijayakumar@ed.ac.uk

capability at all. If treating the walking motion generation as a footstep optimization problem, different formulations exist based on the model that has been used. Considering the fully-actuated LIPM, automatic footstep placement [7] is proposed to simultaneously optimize footstep placements and ZMP trajectories. Formulations considering under-actuated LIPM are also proposed [8] [9]. In those formulations, only footstep locations are optimized. These works introduce a similarity regularization term to penalize the deviation of the optimized footsteps from the referenced ones. In our previous work [10], we replaced the absolute similarity minimization term with a relative one which removes the requirement of the reference footsteps generation plan and makes the footstep optimization truly automatic. In this paper, we adopt the same formulation for lateral stepping motion generation.

For wheeled-legged systems, kinematic motion planning has been demonstrated in [11] [12] [13]. These robots use the legs only as an active suspension system while driving around. Furthermore, the authors impose a quasi-static assumption which limits the type of motion the robot is able to achieve. The work presented in [14] and [15] exploits the full robot dynamics of a wheeled humanoid robot which allows for generating joint torque commands and to achieve compliant interaction. However, the large mobile base of the robot makes it difficult to deal with obstacles or uneven terrain.

More recent results on the wheeled quadrupedal robot ANYmal demonstrate robust dynamic hybrid locomotion capability. The authors of [16] [17] [18] proposed different trajectory optimization (TO) formulations. These formulations integrate the wheels into the control framework so that the robot is capable of performing walking and driving simultaneously.

For wheeled humanoid robots, dynamic balance has been considered by [4] [19] [20]. The common morphology of these wheeled humanoid robots is that the two wheels are directly connected to the same base link. This limits these robots to only use the wheels for driving. In contrast to this, the Ascento [21] robot from ETH and the Handle robot [22] from Boston Dynamics have wheels attached to the leg structure. This makes it possible to potentially use the wheeled leg for walking motions. To overcome obstacles the authors implemented jumping motions. Although this approach is effective, it is not always efficient to avoid obstacles by jumping, i.e. when the obstacle only blocks part of the way of the robot. In this case, the robot can step over the obstacle. This motion essentially requires the robot to be able to balance while it has only one leg on the ground. In this paper, we will demonstrate how this can be achieved through our proposed motion synthesis approach.

### B. Contribution

- We propose to combine Cart-LIPM and under-actuated LIPM to generate hybrid motions for wheeled biped robots. To the best of our knowledge, we are the first to apply Cart-LIPM for rolling motion generation.
- We propose a two degree of freedoms ankle joint (roll-pitch) configuration. This enables the decoupling of

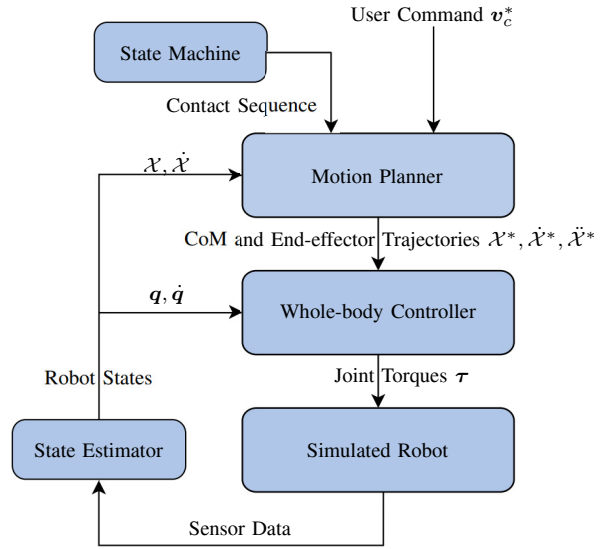

Fig. 2: Control Framework.

rolling and walking motions. It differentiates this work from many existing wheeled legged robots with only one pitch joint in the ankle, such as the wheeled ANYmal [16], Ascento [21] and Handle [22].
- In the whole-body controller, the dynamic nonholonomic constraints on the wheel are defined with respect to the center of the wheel instead of the contact point which gets rid of the extra wheel orientation parametrization as needed in [16].

## II. CONTROL FRAMEWORK

Our control framework takes a simple user command as input and automatically generates the whole-body motion for the robot, including the wheels. The framework has a hierarchical structure as shown in Fig. 2. The inputs are the center of mass (CoM) reference velocity $v_c^* = [\dot{x}_c^*, \dot{y}_c^*, 0]^T$, and the contact sequence generated by the state machine. The contact sequence consists of multiple contact states which include LS (Left-Support), RS (Right-Support) and DS (Double-Support). For rolling motion, we only consider the DS state. For walking and hybrid motions, the contact sequence are the same as shown in Fig. 3. Given the desired CoM velocity and the contact sequence, the motion planner will generate the CoM and end-effector trajectories in Cartesian space. A whole-body controller is then used to track these trajectories by finding the optimal joint torques while considering a set of constraints. Calculated joint torque commands are sent to the simulated robot and all sensor readings have been collected. The state estimator estimates all required robot states based on the raw sensor data and feeds back to the planner and the whole-body controller for the next loop calculation.

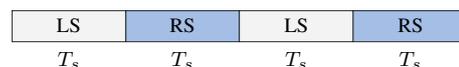

Fig. 3: Walking and hybrid locomotion contact sequence. $T_s$ is the step duration.

In order to fill the gap between the models used in different hierarchies (simple model in the motion planner and full model in the whole-body controller), model predictive control (MPC) is introduced. Reference trajectory re-planning is carried out with a defined MPC update frequency in the motion planner while considering current robot states. This makes the robot more robust and adaptive to external disturbances and internal sensor noise.

## III. MOTION PLANNER

In this section, we will introduce the motion planner. The input to this block is the linear reference velocity for the CoM. No steering control has been assumed to simplify the problem. Before talking about the problem formulation, we will start with the models.

### A. Hybrid Model

Wheeled robots and legged robots are often treated separately due to the different contact natures of rolling motion and walking motion. Rolling motion assumes continuous contact with the ground while walking motion relies on discrete contact changes. These two types of contacts can happen on the same wheel that is rolling on the ground with non-holonomic constraints assumed on it. In its forward rolling direction, the contact position changes along with the rotation of the wheel. In its lateral direction, the contact position stays the same.

To simplify the analysis, we assume that the robot does not steer in the inertial frame and the wheel plane is always parallel to the $x$-$z$ plane of the inertial frame. With this assumption, we can decouple the motion in sagittal plane and frontal plane. Then we propose to use the Cart-LIPM model for sagittal plane rolling motion generation and the under-actuated LIPM for frontal plane walking motion generation as shown in Fig. 4. These two models are essentially different variations of the standard LIPM, which makes it very convenient to compose them. Their dynamics take a similar form:

$$\ddot{x}_c = \omega^2(x_c - x_p) \quad (1)$$

$$\ddot{y}_c = \omega^2(y_c - y_p) \quad (2)$$

where $x_c, y_c$ refers to CoM positions, $x_p, y_p$ refers to ZMP positions, $w = \sqrt{g/z_c}$ is the pendulum frequency with $g$ the gravitational acceleration and $z_c$ the constant CoM height. The key difference between them is the ZMP property: $x_p$ takes continuous values and $y_p$ takes discrete values.

### B. Rolling Motion Planning in the Sagittal Plane

The dynamics of Cart-LIPM (1) gives the instantaneous relationship between CoM and ZMP: the acceleration of CoM $\ddot{x}_c$ is proportional to the distance between it and the ZMP. Therefore, the CoM dynamics can be modulated by controlling the ZMP position $x_p$. Any higher order derivative of $x_p$ can be chosen as the control input, e.g. $\dot{x}_p$ or $\ddot{x}_p$ depending on the actuation type of the wheel. Here, it is assumed that all joints of the robot are torque controlled, the acceleration of the ZMP $\ddot{x}_p$ has been chosen as the control input $u_x$ and the state space model is:

$$\dot{x} = A_x x + B_x u_x \quad (3)$$

$$\begin{bmatrix}\dot{x}_c \\ \ddot{x}_c \\ \dot{x}_p \\ \ddot{x}_p\end{bmatrix} = \begin{bmatrix}0 & 1 & 0 & 0 \\ \omega^2 & 0 & -\omega^2 & 0 \\ 0 & 0 & 0 & 1 \\ 0 & 0 & 0 & 0\end{bmatrix}\begin{bmatrix}x_c \\ \dot{x}_c \\ x_p \\ \dot{x}_p\end{bmatrix} + \begin{bmatrix}0 \\ 0 \\ 0 \\ 1\end{bmatrix}\ddot{x}_p \quad (4)$$

where the state $x = [x_c, \dot{x}_c, x_p, \dot{x}_p]^T$ collects the positions and velocities of CoM and ZMP.

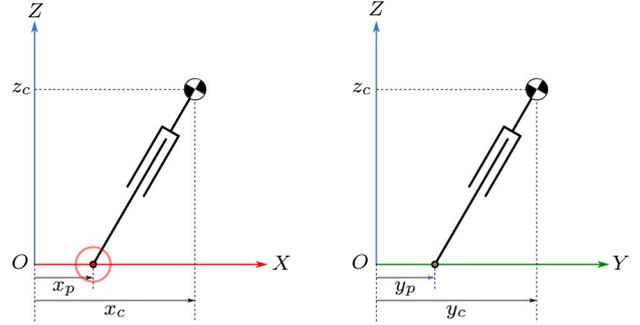

Fig. 4: Cart-LIPM and Under-actuated LIPM.

Since the system is linear, we can apply a linear quadratic regulator (LQR) to stabilize the system. We then compute an optimal state-feedback control for this continuous-time system by minimizing the following quadratic cost function:

$$\begin{aligned}\arg\min_{u_x} &\int_0^\infty x^T Q x + R u_x^2 \\ \text{s.t.} & \\ &\dot{x} = A_x x + B_x u_x\end{aligned} \quad (5)$$

where $Q \in \mathbb{R}^{4\times 4}$ and $R \in \mathbb{R}^{1\times 1}$ are weight matrices corresponding to states and inputs. A diagonal weight matrix $Q$ is chosen and each individual component is selected based on the importance of the corresponding state. Since we are more interested in tracking the reference CoM velocity, we will put higher weight on the CoM velocity state. The input weight matrix $R$ penalizes the control input $\ddot{x}_p$. The wheel acceleration is actually related to the ZMP acceleration $\ddot{x}_p$, so $R$ is indirectly penalizing the wheel actuation. After having $Q$ and $R$ determined, the optimal gain matrix $K$ can be calculated from the associated algebraic Riccati equation [23]. The feedback law which incorporates the reference velocity is:

$$u_x = -K(\hat{x} - x^*) \quad (6)$$

where $\hat{x}$ is the estimated state provided by the state estimator, $x^*$ is the reference state to be tracked. In our case, only the reference CoM velocity $\dot{x}_c^*$ is included in $x^*$.

The optimal control problem is solved with LQR but without considering any hard constraints. Sometimes it is necessary to incorporate system constraints in the planning, such as kinematic limits and friction constraints. Although the template model is very simple, these constraints can be enforced effectively. Then, the problem can be reformulated as a quadratic programming (QP) problem.

*C. Walking Motion Planning in the Frontal Plane*

In the previous section, we addressed how the Cart-LIPM can be used to generate the rolling motion. In this section, we will give details about how the lateral direction walking motion can be achieved. The under-actuated LIPM is used as the template model since the contact between the wheel and the ground is ideally a point. Its dynamics given in equation (2) can be solved analytically:

$$\boldsymbol{y}(t) = \boldsymbol{A}_y(t)\boldsymbol{y}_0 + \boldsymbol{B}_y(t)y_p \quad (7)$$

where $\boldsymbol{A}(t)$ and $\boldsymbol{B}(t)$ are time dependent matrices:

$$\boldsymbol{A}_y(t) = \begin{bmatrix} \cosh(\omega t) & \omega^{-1}\sinh(\omega t) \\ \omega\sinh(\omega t) & \cosh(\omega t) \end{bmatrix}$$
$$\boldsymbol{B}_y(t) = \begin{bmatrix} 1 - \cosh(\omega t) \\ -\omega\sinh(\omega t) \end{bmatrix} \quad (8)$$

It is also the natural dynamics of the model, i.e., the CoM state $\boldsymbol{y}(t) = [y_c(t)\ \dot{y}_c(t)]^T$ evaluates as a function of time for a given initial state $\boldsymbol{y}_0 = [y_c(0)\ \dot{y}_c(0)]^T$ and a fixed support foot placement $y_p$.

Due to the under-actuation of this model, the walking motion planning problem becomes a foot placement optimization problem. Once future steps have been decided, the CoM motion is fixed accordingly. Specifically, given current estimation of the CoM state $\hat{\boldsymbol{y}}_0$ and support foot placement $\hat{y}_{p,0}$, the touchdown moment CoM state can be calculated as:

$$\boldsymbol{y}_1 = \boldsymbol{A}_y(T_0)\hat{\boldsymbol{y}}_0 + \boldsymbol{B}_y(T_0)\hat{y}_{p,0} \quad (9)$$

where $T_0$ is the remaining duration of the current step. The state $\boldsymbol{y}_1$ can not be modified due to fixed $\hat{y}_{p,0}$. Without double support, the robot switches support instantaneously and the final state of the current step becomes the initial state of the next step. After one more step, the CoM state becomes:

$$\boldsymbol{y}_2 = \boldsymbol{A}_y(T_s)\boldsymbol{y}_1 + \boldsymbol{B}_y(T_s)y_{p,1} \quad (10)$$

where $T_s$ is the fixed step duration, the state $\boldsymbol{y}_2$ is only related to foot placement $y_{p,1}$. This process can be repeated:

$$\boldsymbol{y}_3 = \boldsymbol{A}_y(T_s)\boldsymbol{y}_2 + \boldsymbol{B}_y(T_s)y_{p,2}$$
$$\ldots \quad (11)$$
$$\boldsymbol{y}_{N+1} = \boldsymbol{A}_y(T_s)\boldsymbol{y}_N + \boldsymbol{B}_y(T_s)y_{p,N}$$

Where $N$ is the number of steps to be optimized. It can be concluded that all the future CoM states are a function of future steps. In order to optimize future CoM states, we could collect all future step locations as the optimization variable $\boldsymbol{y}_p = [y_{p,1}\ y_{p,2}\ \ldots\ y_{p,N}]^T$. The primary goal in our case is to track the reference CoM velocity $v_y^*$. The other important task is the previously proposed relative distance similarity regularization [10], which keeps the feet away from each other to avoid self-collision. The overall cost is defined as:

$$\arg\min_{\boldsymbol{y}_p} \sum_{i=1}^{N} Q(\dot{y}_{c,i+1} - v_y^*)^2 + R(\Delta y_{p,i} - \Delta y_{p,i}^*)^2 \quad (12)$$

where $\Delta y_{p,i} = y_{p,i} - y_{p,i-1}$ is the step length between two adjacent steps, $\Delta y_{p,i}^*$ is the desired step length defined as $\Delta y_{p,i}^* = s \cdot (-1)^i d$, where $d$ is the desired inter-feet clearance, and $s$ indicates the current support phase (1 for left support and -1 for right support). The step length similarity cost term only encourages the feet stay away from each other. In extreme scenarios such as when the robot has been heavily disturbed in the lateral direction, hard constraints on the step length are necessary to prevent feet self-collision or leg over stretching:
$d_{min} < s \cdot (-1)^i \Delta p_{y,i} < d_{max} \quad (i = 1, ..., N)$.

It is worth mentioning that only the first optimal step position $y_{p,1}^*$ will be used to adapt the swing foot trajectory based on the MPC implementation. The CoM trajectory can be calculated from (7). The updated CoM trajectory and swing foot trajectory are sent to the whole-body controller to track.

IV. WHOLE-BODY CONTROLLER

The whole-body controller is used to track trajectories given from the motion planner while satisfying specified constraints. In the case of hybrid locomotion, the targets which need to be tracked are CoM, support wheel center position and swing wheel center position. The wheels need to be coordinated with all other joints to achieve these tracking tasks. The full dynamic model of the multi-rigid-body system is thus considered. The equation of motion is:

$$\boldsymbol{M}(\boldsymbol{q})\ddot{\boldsymbol{q}} + \boldsymbol{h}(\boldsymbol{q},\dot{\boldsymbol{q}}) = \boldsymbol{S}^T\boldsymbol{\tau} + \boldsymbol{J}_C(\boldsymbol{q})^T\boldsymbol{\lambda} \quad (13)$$

where $\boldsymbol{M}(\boldsymbol{q}) \in \mathbb{R}^{(n+6)\times(n+6)}$, $\boldsymbol{h}(\boldsymbol{q},\dot{\boldsymbol{q}}) \in \mathbb{R}^{n+6}$ are the mass matrix and nonlinear term, $\boldsymbol{q} \in SE(3) \times \mathbb{R}^n$ represents the configuration of the system which includes the pose of the base link and joint positions of $n$ actuated joints, and $\dot{\boldsymbol{q}} \in \mathbb{R}^{n+6}$ and $\ddot{\boldsymbol{q}} \in \mathbb{R}^{n+6}$ are the generalized velocity and acceleration. The selection matrix $\boldsymbol{S} = [\boldsymbol{0}_{n\times 6}\ \boldsymbol{I}_{n\times n}]$ selects the actuated joints. $\boldsymbol{\tau} \in \mathbb{R}^n$ is the actuated joint torques. $\boldsymbol{J}_C \in \mathbb{R}^{(3n_c)\times(n+6)}$ is a concatenated contact Jacobian $\boldsymbol{J}_C = [\boldsymbol{J}_1^T\ \boldsymbol{J}_2^T\ \ldots\ \boldsymbol{J}_{n_c}^T]^T$ and $n_c$ is the number of contacts. $\boldsymbol{\lambda} \in \mathbb{R}^{3n_c}$ is the concatenated contact forces corresponding to the contact Jacobian.

*A. Optimization Formulation*

The goal of the optimization is to track a set of tasks. When there is not enough solution space to realise all tasks at the same time, a mechanism is needed to resolve the conflicts between tasks. There are two choices: a weighted approach or a strict hierarchy approach. In fact, the two can be combined in a general formulation which imposes strict priorities between different hierarchies while allowing soft compromises among tasks within the same hierarchy. In particular, the whole-body control problem is formulated as a cascade of quadratic programming (QP) problems which are solved in a strict prioritized order [24]. The optimization variable is $\boldsymbol{\xi} = [\ddot{\boldsymbol{q}}^T \ \boldsymbol{\lambda}^T]^T$. A task with priority $p$ is defined as:

$$T_p : \begin{cases} \boldsymbol{W}_{eq,p} \left( \boldsymbol{A}_p \boldsymbol{\xi} - \boldsymbol{b}_p \right) = \boldsymbol{0} \\ \boldsymbol{W}_{ineq,p} \left( \boldsymbol{C}_p \boldsymbol{\xi} - \boldsymbol{d}_p \right) \leq \boldsymbol{0} \end{cases} \quad (14)$$

where $\boldsymbol{A}_p$ and $\boldsymbol{b}_p$ defines the equality constraints and $\boldsymbol{C}_p$ and $\boldsymbol{d}_p$ defines the inequality constraints, they are concatenated from all tasks with the same priority $p$. $\boldsymbol{W}_{eq,p}$ and $\boldsymbol{W}_{ineq,p}$ are diagonal weight matrices that weigh tasks in this hierarchy.

### B. Tasks

The previous section gives the general optimization formulation (14). In this section, we will highlight tasks that are important for the hybrid locomotion.

*1) Dynamic constraints and torque limits:* In fact, the system dynamic equation (13) can be rewritten in two parts [25]:

$$\boldsymbol{M}_\lambda \ddot{\boldsymbol{q}} + \boldsymbol{h}_\lambda = \boldsymbol{J}_{C,\lambda}^T \boldsymbol{\lambda} \quad (15a)$$
$$\boldsymbol{M}_\tau \ddot{\boldsymbol{q}} + \boldsymbol{h}_\tau = \boldsymbol{\tau} + \boldsymbol{J}_{C,\tau}^T \boldsymbol{\lambda} \quad (15b)$$

where the upper six rows corresponds to the floating base and are used as system dynamic constraints. The lower actuated part can be used to calculate joint torques from joint acceleration and contact forces: $\boldsymbol{\tau} = \boldsymbol{M}_\tau \ddot{\boldsymbol{q}} + \boldsymbol{h}_\tau - \boldsymbol{J}_{C,\tau}^T \boldsymbol{\lambda}$. Additionally, torque limit constraints $\boldsymbol{\tau} \in [\boldsymbol{\tau}^-, \boldsymbol{\tau}^+]$ can be defined with it.

*2) Friction constraints and unilateral constraints:* The friction cone is approximated with a friction pyramid and then enforced on the contact forces $\boldsymbol{\lambda}$. In the local frame of each contact force $\boldsymbol{f}_i$, the constraints can be written as: $|f_{i,x}| \leq \mu f_{i,z}$, $|f_{i,y}| \leq \mu f_{i,z}$ ($\mu$: friction coefficient). The unilateral constraints requires the $z$ component of the contact force to be positive $f_{i,z} > 0$.

*3) Nonholonomic constraints:* Nonholonomic constraints ensure pure rolling for the wheel which is in contact with the ground (no slipping in wheel radial direction and no sliding in wheel axial direction). Fig. 5 shows our two DoF ankle joint configuration. Both roll joint and pitch joint are actively controlled. The roll joint moves in a limited range while the pitch joint can rotate continuously without limits.

Instead of developing the nonholonomic constraints on the contact point $C$, we derive it with respect to the center of the wheel $W$. In the inertia frame $\mathcal{I}$, the velocity of point $C$ can be related to the velocity of the point $W$ through:

$$\boldsymbol{v}_C = \boldsymbol{v}_W + \boldsymbol{\omega}_W \times \boldsymbol{r}_{WC} \quad (16)$$

where $\boldsymbol{v}_W = \boldsymbol{J}_W \dot{\boldsymbol{q}}$ is linear velocity of the center of the wheel (it is not related to the wheel rotation since $W$ locates on its rotation axis). $\boldsymbol{\omega}_W$ is the rotational velocity of the wheel and it can be expressed as $\boldsymbol{\omega}_W = \hat{\boldsymbol{\omega}}_W \cdot ||\hat{\boldsymbol{\omega}}_W|| = \hat{\boldsymbol{y}}_W \cdot \dot{q}_W$, where $\dot{q}_W$ refers to the joint velocity of the wheel. $\boldsymbol{r}_{WC}$ is the vector pointing from $W$ to $C$. Kinematic nonholonomic constraints require the velocity of the contact point $C$ to be equal to zero with respect to the ground, $\boldsymbol{v}_C = 0$. Combining this with (16), the constraint becomes:

$$\boldsymbol{v}_W + \boldsymbol{\omega}_W \times \boldsymbol{r}_{WC} = 0 \quad (17)$$

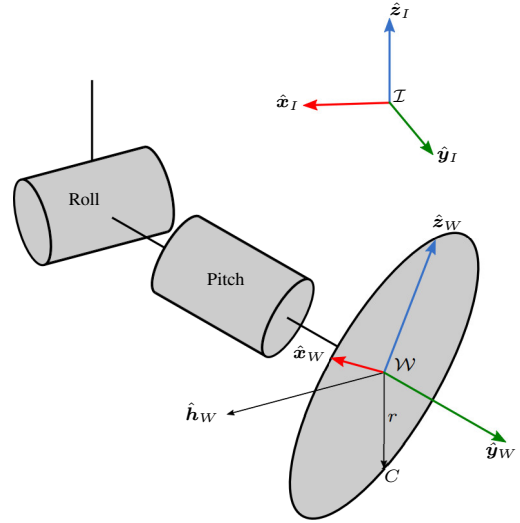

Fig. 5: The two DoF ankle joints consists of a roll joint and a pitch joint. $I$ is the origin of the inertia frame $\mathcal{I}$. $W$ stands for the center of the wheel and it is also the origin of the wheel link frame $\mathcal{W}$. $C$ is the contact point between the wheel and the ground. $\hat{\boldsymbol{h}}_W$ indicates the heading direction of the wheel and it is defined from the wheel rotation axis and the ground norm axis $\hat{\boldsymbol{h}}_W = \hat{\boldsymbol{y}}_W \times \hat{\boldsymbol{z}}_I$. $r$ is the radius of the wheel.

Differentiating with respect to time gives the acceleration level nonholonomic constraints:

$$\dot{\boldsymbol{v}}_W + \dot{\boldsymbol{\omega}}_W \times \boldsymbol{r}_{WC} + \boldsymbol{\omega}_W \times (\boldsymbol{\omega}_W \times \boldsymbol{r}_{WC}) = 0 \quad (18)$$

where $\dot{\boldsymbol{v}}_W = \boldsymbol{J}_W \ddot{\boldsymbol{q}} + \dot{\boldsymbol{J}}_W \dot{\boldsymbol{q}}$ is the linear acceleration of $W$, $\dot{\boldsymbol{\omega}}_W$ is the rotational acceleration of the wheel and is related to the wheel joint acceleration $\ddot{q}_W$ through $\dot{\boldsymbol{\omega}}_W = \hat{\boldsymbol{y}}_W \cdot \ddot{q}_W$. Substituting these relations into equation (18):

$$\boldsymbol{J}_W \ddot{\boldsymbol{q}} + \dot{\boldsymbol{J}}_W \dot{\boldsymbol{q}} + [\hat{\boldsymbol{y}}_W]_\times \boldsymbol{r}_{WC} \ddot{q}_W + [\hat{\boldsymbol{y}}_W]_\times^2 \boldsymbol{r}_{WC} \dot{q}_W^2 = 0 \quad (19)$$

where $[\ ]_\times$ is the skew-symmetric cross product operator, A wheel selection matrix $\boldsymbol{S}_W$ can be defined to select out the wheel joint velocity and acceleration: $\dot{q}_W = \boldsymbol{S}_W \dot{\boldsymbol{q}}$, $\ddot{q}_W = \boldsymbol{S}_W \ddot{\boldsymbol{q}}$. Finally, equation (19) becomes:

$$(\boldsymbol{J}_W + [\hat{\boldsymbol{y}}_W]_\times \boldsymbol{r}_{WC} \boldsymbol{S}_W) \ddot{\boldsymbol{q}} = -\dot{\boldsymbol{J}}_W \dot{\boldsymbol{q}} - [\hat{\boldsymbol{y}}_W]_\times^2 \boldsymbol{r}_{WC} (\boldsymbol{S}_W \dot{\boldsymbol{q}})^2 \quad (20)$$

This gives the dynamic nonholonomic constraints for the wheel rolling on the ground.

Although the nonholonomic constraints are defined with respect to the point $W$, the location of the contact point $C$ is needed since $\boldsymbol{r}_{WC}$ has been used in previous derivation. The contact point is on the edge of the wheel, so the distance from $W$ to $C$ is a known constant $r$. Since $\boldsymbol{r}_{WC} = \hat{\boldsymbol{r}}_{WC} r$, we still need to find out its unit direction vector $\hat{\boldsymbol{r}}_{WC}$. The heading direction of the wheel is defined from the wheel axis $\hat{\boldsymbol{y}}_W$ and the ground norm $\hat{\boldsymbol{z}}_I$: $\hat{\boldsymbol{h}}_W = \hat{\boldsymbol{y}}_W \times \hat{\boldsymbol{z}}_I$. Then the direction vector $\hat{\boldsymbol{r}}_{WC}$ can be calculated: $\hat{\boldsymbol{r}}_{WC} = \hat{\boldsymbol{y}}_W \times \hat{\boldsymbol{h}}_W = \hat{\boldsymbol{y}}_W \times (\hat{\boldsymbol{y}}_W \times \hat{\boldsymbol{z}}_I)$.

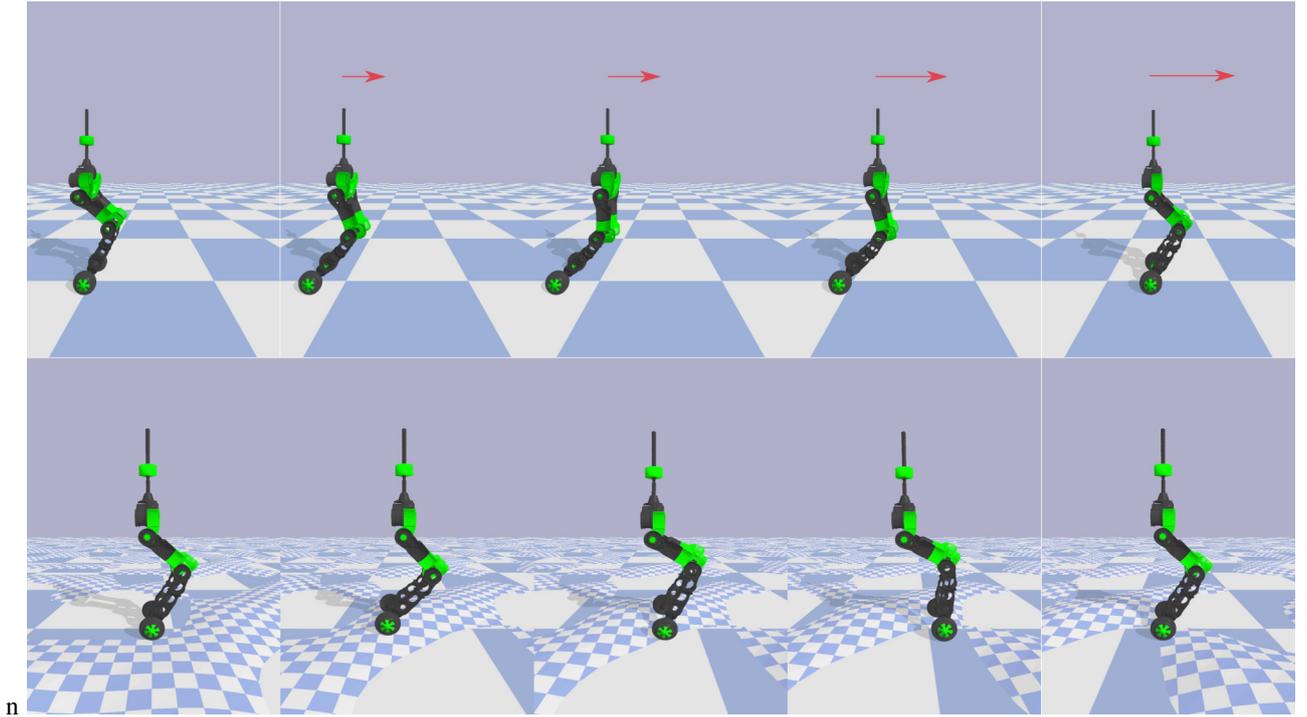

Fig. 6: Rolling motion on flat ground (top row) and uneven terrain (bottom row).

*4) Cartesian space motion tracking:* The outputs of the motion planner are Cartesian space trajectories generated based on the reduced order template models. The whole-body controller needs to track them as close as possible. The commanded acceleration is defined based on the desired trajectories:

$$\ddot{\mathcal{X}}_c = \boldsymbol{K}_P \boldsymbol{E}(\mathcal{X}^*, \mathcal{X}) + \boldsymbol{K}_D(\dot{\mathcal{X}}^* - \dot{\mathcal{X}}) + \ddot{\mathcal{X}}^* \quad (21)$$

where $\mathcal{X}^*, \dot{\mathcal{X}}^*, \ddot{\mathcal{X}}^*$ are the desired pose, velocity and acceleration. $\mathcal{X}, \dot{\mathcal{X}}$ are the current pose and velocity of the controlled frame. $\boldsymbol{E}(\mathcal{X}^*, \mathcal{X})$ gives the error between two poses in $SE(3)$. $\boldsymbol{K}_P, \boldsymbol{K}_D$ are feedback gain matrices. The commanded task space acceleration is related to the joint space acceleration:

$$\ddot{\mathcal{X}}_c = \boldsymbol{J}_T \ddot{\boldsymbol{q}} + \dot{\boldsymbol{J}}_T \dot{\boldsymbol{q}} \quad (22)$$

where $\boldsymbol{J}_T$ refers to the task jacobian matrix. For the centroidal task, it is the centroidal momentum matrix. For the swing foot tracking, it becomes the swing foot jacobian matrix.

## V. SIMULATION

Given the motion planner and whole-body control, we are ready to generate motions for the wheeled biped robot. Rolling and walking are first generated separately based on their respective template models. Hybrid locomotion is then performed to validate the composition of the two. Two scenarios have been designed to show the usefulness of the hybrid locomotion mode. The simulation is conducted in *PyBullet* [26], a Python module that extends the *Bullet* physics engine. The robot used here is the lower-body of the humanoid robot *COMAN+* [27] with attached wheels.

### A. Rolling

Rolling is the most basic locomotion mode of the wheeled biped robot and it refers to the motion that the robot rolls on its wheels. Both wheels stay on the ground throughout the whole motion. In other words, the robot only has a double support phase. The obvious benefit of not having single support is that the robot does not need to handle lateral balancing. The drawback is that the robot can only navigate over relatively regular terrain. When encountering significant obstacles, the robot has to re-plan its route and drive around.

The simulated motion is demonstrated in the top of the Fig. 6. The snapshots shows a generated rolling motion that starts from zero velocity and accelerates to a given desired velocity $1.0m/s$. Here, the weights $\boldsymbol{Q} = \text{diag}([1\ 10^4\ 1\ 10^3])$ and $\boldsymbol{R} = 10$ have been tuned to make the robot perform more aggressively. This can be seen from the third picture of the top row in which the robot almost fully extend its legs to maximize the CoM acceleration. A detailed plot of CoM velocities in forward direction is given in Fig. 7.

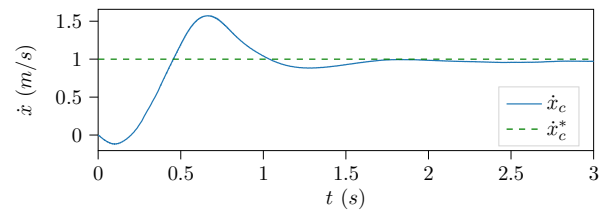

Fig. 7: Rolling motion CoM velocity trajectory (blue). The green dashed line stands for the desired velocity that has been given to the planner.

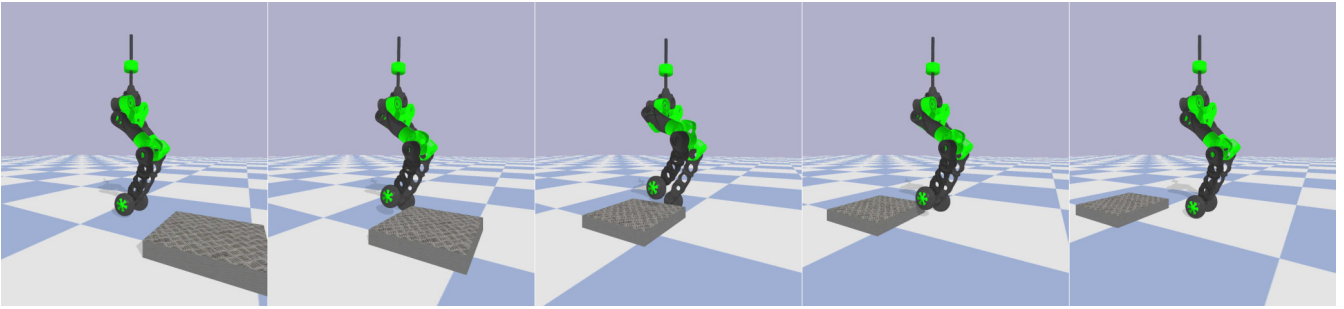

Fig. 8: Wheeled biped robot goes over obstacle. The size of the obstacle is $0.5{\times}0.5{\times}0.08m$ and the radius of the wheel is $0.08m$. The current speed of the robot is around $4.5m/s$.

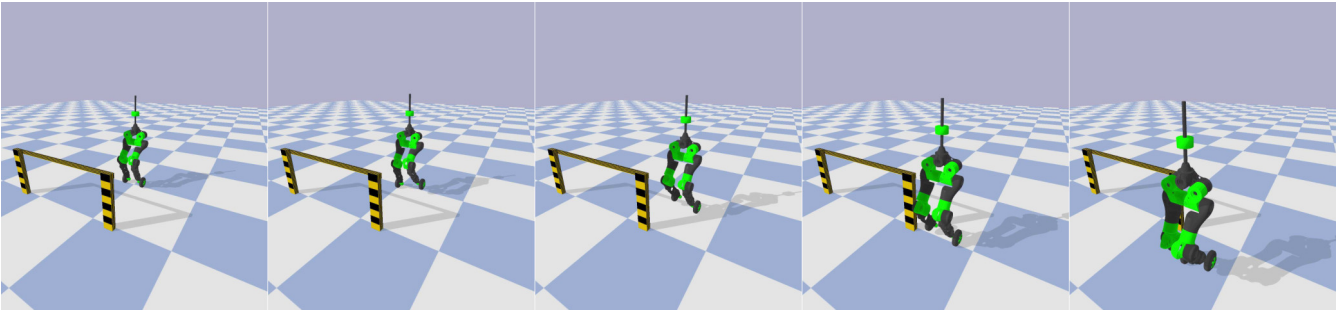

Fig. 9: Wheeled biped robot goes around barrier without steering.

To further test the robustness of the control system, we repeated the same motion with uneven terrain added to the scene as shown in the bottom of the Fig. 6. The terrain is generated with Perlin noise with the maximum height of $0.08m$. The robot has no knowledge of the terrain but is still able to traverse.

### B. Walking

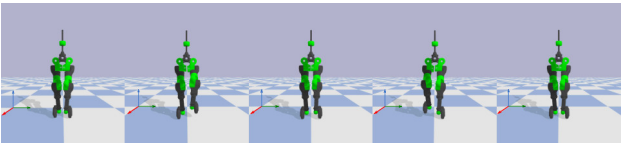

Fig. 10: Walking motion.

Walking mode is another basic locomotion mode of the robot. The walking mode described here is a mode in which both wheels of the robot are treated as point feet and the robot steps on them. Both wheels do not rotate continuously in this mode. Besides that, only single support has been considered for walking as suggested in section III-C. Fig. 10 shows the walking motion. Despite the fact that we are demonstrating motion in the lateral direction only, we do need to take care of balance in the forward direction since the robot can only get point support from the wheel. To achieve 3D walking motion, we have composed two under-actuated LIPMs in both directions, more details can be found in our previous work [10]. For walking in place, the desired velocities in both directions are simply set to zero.

### C. Hybrid Locomotion

Hybrid locomotion mode combines rolling and walking. The contact sequence for hybrid locomotion is the same as walking as shown in Fig. 3. The wheel in contact with ground is rolling in the forward direction while pushing the robot in lateral direction. The wheel in the air is tracking a trajectory defined for its center, while its rotation is subject to a minimum torque task. Two scenarios have been designed to show the usefulness this mode.

In the first scenario, an obstacle is presented in front of the robot and blocks half of the robot, but the robot is able to step over it while the other wheel on the ground still keeps rolling. The process of going over the obstacle is plotted in Fig. 8. The height of the obstacle is the same as the radius of the wheel and the width and length are both 0.5m. It is very challenging either to roll over (due to the height) or to walk over (due to the length). However it can be achieved through the enabled hybrid locomotion; what is more, the "step length" in hybrid mode is proportional to the speed of the robot.

In the second scenario, a much higher barrier is placed in front of the robot. Instead of steering away to move around it, with the hybrid mode, the robot can step aside while rolling forward. This movement is shown in Fig. 9.

The above simulations show that the hybrid locomotion mode indeed provides the robot more possibilities to traverse cluttered environments. Another possible but not demonstrated scenario is for the robot to step over a small trench on the ground. To summarise, the hybrid locomotion mode makes the wheeled biped robots closer to the real world environment and can be potentially very useful in our daily lives.

## VI. Conclusion and Future Works

In this paper, we have demonstrated how different types of motion such as rolling, walking and hybrid motions can be generated from our proposed framework. It is realised through the composition of decoupled rolling motion and walking motion. The Cart-LIPM and the under-actuated LIPM are proposed to model these two motions. Due to the model linearity, the motion planning can be executed in real time which enables fast online MPC implementation, significantly increasing the robustness of the motion. At the end, we have demonstrated the usefulness of the enabled hybrid locomotion. However, the proposed composition method has limitations introduced by the template models such as the requirement of constraint CoM height. This prohibits more dynamic motion, such as jumping, which could be very effective in certain circumstances. The motion composition mentioned in this paper only composes the decoupled motions in different directions. Actually, the composition can happen in a single direction. For example in the forward rolling direction, rolling and walking can be also combined. Another issue that has not been mentioned in the paper is the steering control which could be potentially explored, especially in the hybrid locomotion scenario.

## Acknowledgment

This research is supported by the Future AI and Robotics Hub for Space (FAIR-SPACE) funded by the Engineering and Physical Sciences Research Council (EPSRC) (EP/R026092/1). The authors also would like to thank Traiko Dinev, Vladimir Ivan, Matt Timmons-Brown and Daniel Gordon for their suggestions on the writing of the paper.

## References

[1] R. P. M. Chan, K. A. Stol, and C. R. Halkyard, "Review of modelling and control of two-wheeled robots," *Annual reviews in control*, vol. 37, no. 1, pp. 89–103, 2013.

[2] J. Akesson, A. Blomdell, and R. Braun, "Design and control of yaipan inverted pendulum on two wheels robot," in *2006 IEEE Conference on Computer Aided Control System Design, 2006 IEEE International Conference on Control Applications, 2006 IEEE International Symposium on Intelligent Control*, pp. 2178–2183, IEEE, 2006.

[3] S. Jeong and T. Takahashi, "Wheeled inverted pendulum type assistant robot: design concept and mobile control," *Intelligent Service Robotics*, vol. 1, no. 4, pp. 313–320, 2008.

[4] M. Zafar, S. Hutchinson, and E. A. Theodorou, "Hierarchical optimization for whole-body control of wheeled inverted pendulum humanoids," in *2019 International Conference on Robotics and Automation (ICRA)*, pp. 7535–7542, IEEE, 2019.

[5] S. Kajita, F. Kanehiro, K. Kaneko, K. Fujiwara, K. Harada, K. Yokoi, and H. Hirukawa, "Biped walking pattern generation by using preview control of zero-moment point," in *IEEE International Conference on Robotics and Automation*, vol. 2, pp. 1620–1626, IEEE.

[6] P.-B. Wieber, "Trajectory free linear model predictive control for stable walking in the presence of strong perturbations," in *IEEE-RAS International Conference on Humanoid Robots*.

[7] A. Herdt, H. Diedam, P.-B. Wieber, D. Dimitrov, K. Mombaur, and M. Diehl, "Online walking motion generation with automatic footstep placement," *Advanced Robotics*, vol. 24, no. 5-6, pp. 719–737, 2010.

[8] S. Faraji, S. Pouya, C. G. Atkeson, and A. J. Ijspeert, "Versatile and robust 3d walking with a simulated humanoid robot (atlas): A model predictive control approach," in *IEEE International Conference on Robotics and Automation*, pp. 1943–1950.

[9] S. Feng, X. Xinjilefu, C. G. Atkeson, and J. Kim, "Robust dynamic walking using online foot step optimization," pp. 5373–5378.

[10] S. Xin, R. Orsolino, and N. Tsagarakis, "Online relative footstep optimization for legged robots dynamic walking using discrete-time model predictive control," in *2019 IEEE/RSJ International Conference on Intelligent Robots and Systems*, IEEE, 2019.

[11] C. Grand, F. Benamar, and F. Plumet, "Motion kinematics analysis of wheeled–legged rover over 3d surface with posture adaptation," *Mechanism and Machine Theory*, vol. 45, no. 3, pp. 477–495, 2010.

[12] K. Nagano and Y. Fujimoto, "The stable wheeled locomotion in low speed region for a wheel-legged mobile robot," in *2015 IEEE International Conference on Mechatronics (ICM)*, pp. 404–409, IEEE, 2015.

[13] M. Kamedula, N. Kashiri, and N. G. Tsagarakis, "On the kinematics of wheeled motion control of a hybrid wheeled-legged centauro robot," in *2018 IEEE/RSJ International Conference on Intelligent Robots and Systems (IROS)*, pp. 2426–2433, IEEE, 2018.

[14] L. Sentis, J. Petersen, and R. Philippsen, "Implementation and stability analysis of prioritized whole-body compliant controllers on a wheeled humanoid robot in uneven terrains," *Autonomous Robots*, vol. 35, no. 4, pp. 301–319, 2013.

[15] A. Dietrich, K. Bussmann, F. Petit, P. Kotyczka, C. Ott, B. Lohmann, and A. Albu-Schäffer, "Whole-body impedance control of wheeled mobile manipulators," *Autonomous Robots*, vol. 40, no. 3, pp. 505–517, 2016.

[16] M. Bjelonic, C. D. Bellicoso, Y. De Viragh, D. Sako, F. D. Tresoldi, F. Jenelten, and M. Hutter, "Keep Rollin'-Whole-Body Motion Control and Planning for Wheeled Quadrupedal Robots," vol. 4, no. 2, pp. 2116–2123.

[17] Y. de Viragh, M. Bjelonic, C. D. Bellicoso, F. Jenelten, and M. Hutter, "Trajectory optimization for wheeled-legged quadrupedal robots using linearized zmp constraints," *IEEE Robotics and Automation Letters*, vol. 4, no. 2, pp. 1633–1640, 2019.

[18] M. Bjelonic, P. K. Sankar, C. D. Bellicoso, H. Vallery, and M. Hutter, "Rolling in the deep–hybrid locomotion for wheeled-legged robots using online trajectory optimization," *arXiv preprint arXiv:1909.07193*, 2019.

[19] M. Stilman, J. Olson, and W. Gloss, "Golem Krang: Dynamically stable humanoid robot for mobile manipulation," in *2010 IEEE International Conference on Robotics and Automation*, pp. 3304–3309, IEEE.

[20] G. Zambella, G. Lentini, M. Garabini, G. Grioli, M. G. Catalano, A. Palleschi, L. Pallottino, A. Bicchi, A. Settimi, and D. Caporale, "Dynamic whole-body control of unstable wheeled humanoid robots," *IEEE Robotics and Automation Letters*, vol. 4, no. 4, pp. 3489–3496, 2019.

[21] V. Klemm, A. Morra, C. Salzmann, F. Tschopp, K. Bodie, L. Gulich, N. Küng, D. Mannhart, C. Pfister, M. Vierneisel, *et al.*, "Ascento: A two-wheeled jumping robot," in *2019 International Conference on Robotics and Automation (ICRA)*, pp. 7515–7521, IEEE, 2019.

[22] Boston Dynamics, "Introducing handle." https://youtu.be/-7xvqQeoA8c, 2017.

[23] E. D. Sontag, *Mathematical control theory: deterministic finite dimensional systems*, vol. 6. Springer Science & Business Media, 2013.

[24] C. D. Bellicoso, C. Gehring, J. Hwangbo, P. Fankhauser, and M. Hutter, "Perception-less terrain adaptation through whole body control and hierarchical optimization," in *2016 IEEE-RAS 16th International Conference on Humanoid Robots (Humanoids)*, pp. 558–564, IEEE, 2016.

[25] A. Herzog, L. Righetti, F. Grimminger, P. Pastor, and S. Schaal, "Balancing experiments on a torque-controlled humanoid with hierarchical inverse dynamics," in *2014 IEEE/RSJ International Conference on Intelligent Robots and Systems*, pp. 981–988, IEEE, 2014.

[26] E. Coumans and Y. Bai, "Pybullet, a python module for physics simulation for games, robotics and machine learning," *https://pybullet.org*, 2016.

[27] C. Zhou and N. Tsagarakis, "On the comprehensive kinematics analysis of a humanoid parallel ankle mechanism," *Journal of Mechanisms and Robotics*, vol. 10, no. 5, 2018.